\documentclass{article}

\usepackage[final]{neurips_2021}

\usepackage[utf8]{inputenc} 
\usepackage[T1]{fontenc}    
\usepackage{hyperref}       
\usepackage{url}            
\usepackage{booktabs}       
\usepackage{amsfonts}       
\usepackage{nicefrac}       
\usepackage{microtype}      
\usepackage{xcolor}         
\usepackage{graphicx}
\usepackage{subfig}

\title{Machine Learning Models Evaluation and Feature Importance Analysis on NPL Dataset}

\author{
 Rufael Fekadu \thanks{Work done when the authors were a research intern at Chapa.} \textsuperscript {  $\dagger$} \\
  School of Electrical and Computer Engineering\\
  Addis Ababa Science and Technology University\\
  Addis Ababa, Ethiopia \\
   \\
  \And
 Anteneh Getachew \textsuperscript{$\ast$ $\dagger$} \\
  School of Electrical and Computer Engineering\\
  Addis Ababa Science and Technology University\\
  Addis Ababa, Ethiopia \\
  \\
 \And
 Yishak Tadele \textsuperscript{$\ast$ $\dagger$} \\
  School of Electrical and Computer Engineering\\
  Addis Ababa Science and Technology University\\
  Addis Ababa, Ethiopia \\
   \\
    \AND
   Nuredin Ali \thanks{Equally contributed to this work.} \\
  Chapa AI Research (ChAIR) \\
  Chapa Financial Technologies \\
  Addis Ababa, Ethiopia \\
   \\
\And
 Israel Goytom \\
  Chapa AI Research (ChAIR)\\
  Chapa Financial Technologies\\
  Addis Ababa, Ethiopia \\
   \\

}

\begin{document}

\maketitle
\begin{abstract}
Predicting the probability of non-performing loans for individuals has a vital and beneficial role for banks to decrease credit risk and make the right decisions before giving the loan. The trend to make these decisions are based on credit study and in accordance with generally accepted standards, loan payment history, and demographic data of the clients. In this work, we evaluate how different Machine learning models such as Random Forest, Decision tree, KNN, SVM, and XGBoost perform on the dataset provided by a private bank in Ethiopia. Further, motivated by this evaluation we explore different feature selection methods to state the important features for the bank. Our findings show that XGBoost achieves the highest F1 score on the KMeans SMOTE over-sampled data. We also found that the most important features are the age of the applicant, years of employment, and total income of the applicant rather than collateral-related features in evaluating credit risk. 

\end{abstract}

\section{Introduction}

Loans are one of the primary sources of income for both private and commercial banks. Large portion of banks' profit directly comes from the interests on loans given. Profitable banks play a vital role in the economic development of each country and a bank with a poor management system is an obstacle to the economic development of the country. The role of banks is spiral in developing countries as there is a growing need for financial institutions. 
 
Although most loans are paid back based on their schedule some others default. Those defaulting loans are known as non-performing loans(NPL). In recent years, evaluating and predicting the non-repayment ability of a customer is one of the most challenging issues in commercial banks. \cite{adhikary_nonperforming_nodate} The presence of huge non-performing loans in the banking industry highly affects the safe operation of banks, and may lead to bank failures and even nation-wide financial crisis. Since lending without better evaluation techniques may lead to immediate losses, finding a way to reduce the credit risk of financial institutions has been a major area of concern for many researchers around the globe. 

Banks use different criteria to predict the creditworthiness of applicants. To evaluate the creditworthiness of customers most Ethiopian banks focus on loan payment history, demographic data, and collateral quality of the clients. Although this system tends to work well sometimes loans default due to various reasons. In recent years the attention of Ethiopian based banks on their non-performing loans has grown. Different types of initiatives should be made to reduce the amount of non-performing loans to strengthen the financial institutions. Globally, plenty of research has been conducted \cite{li_comparison_2010} to determine the creditworthiness of a customer, partitioning the credit groups of good or bad payers.

The rise of AI in finance and beyond has understandably garnered a great deal  of attention in recent years \cite{heath_prediction_2019}. Different machine learning techniques have been evaluated to predict non-performing loans on different banks' datasets.  
         
This work aims to evaluate the performance of various machine learning algorithms on the dataset. We also explore different feature selection mechanisms to  select the most important features and state those features to the bank.  To the best of our knowledge, there is no extensive research on identifying defaulting loans in Ethiopia, which results in the problem getting worse.

\section{Related Works and Background}
There have been various works addressing the performance of machine learning models on predicting NPLs in different countries. 
Banks follow different criteria to predict whether a loan will default or not. Various machine learning models have been evaluated and their performance differs based on the data used. 

\cite{zhu_study_2019} Applied different machine learning algorithms for loan prediction. Their work concluded that random forest has much better accuracy than other algorithms such as logistic regression, decision tree, and support vector machine. \cite{ghatasheh_business_2014} Makes a comparative analysis of different algorithms and concludes that random forest is among the best methods to try for credit risk prediction. \cite{breeden_survey_2020} After reviewing various machine learning methods available and the equally numerous applications, it is clear that declaring a single best method is impossible. Methods have specific strengths and weaknesses that align with different applications. A comprehensive study is conducted to compare the performance of XGBoost algorithm with logistic regression \cite{li_credit_2019}. Their results show that the XGBoost algorithm works much better. 


\cite{pradhan_performance_2020} Established a solid comparison between different classification algorithms. As a result, they found random forest as the best performing model, and Naive Bayes as the worst in terms of accuracy and Area Under the Curve (AUC).  \cite{granstrom_loan_2019} Investigated Logistic Regression, Random Forest, Decision Tree, AdaBoost, XGBoost, Artificial Neural Network and Support Vector Machine. They applied SMOTE to overcome the problem of imbalance. Their results shows that XGBoost without SMOTE implementation obtained the best result with respect to the chosen model evaluation metric which is F1-Score. 

The work by \cite{chen_research_2021} used K-means SMOTE to solve the imbalance problem and feature importance scores generated by random forest are fed into BP neural networks as an initial weight. The work concludes that the improved version of smote algorithm (k-means SMOTE) effectively solves the imbalance problem and the introduced technique improves the prediction performance of the model to a certain extent. \cite{gultekin_variable_2018} Evaluated various machine learning algorithms to predict default loans and concluded that Neural Network model was the best model with higher accuracy and low average square error also Random Forest model better resulted than Logistic Regression model.

\section{Dataset and Challenges}
\subsection{Dataset Description}
We used a dataset from one of the largest Ethiopian Private Banks, Bank of Abyssinia. There were two separate files, applicants record and loan record. The application record contains information about the applicants. It has 438557 records with 18 features. The loan record contains the monthly repayment status of each applicant. It has 1048575 records with 3 features. We merged both files based on the Customer ID, and got 36,326 records. The dataset has class imbalance, null entries, invalid features, and duplicate records. 

\subsection{Exploratory Data Analysis(EDA)}

\begin{figure}[h]
  \centering
  \includegraphics[width=0.5\textwidth]{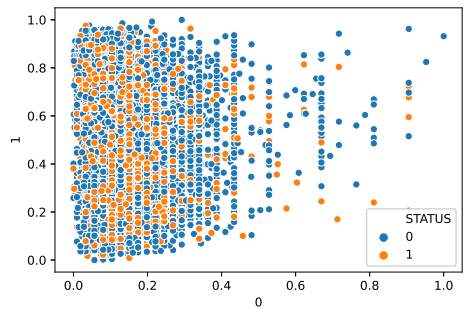}
 \caption{Shows the distribution of the target class.}\label{fig 1:}
\end{figure}

In order to get a better understanding of the dataset, we provided some explanatory plots. The distribution of the target class is shown in Figure \ref{fig 1:}.    
There is a high correlation between \begin{small}CNT\_FAM\_MEMBERS\end{small} and \begin{small}CNT\_CHILDREN\end{small}. Since both features are not correlated to the target class \begin{small}STATUS\end{small}. \begin{small}CNT\_CHILDREN\end{small}. Hence,a new feature called CNT\_ADULT
was created by subtracting CNT\_CHILDREN from CNT\_FAM\_MEMBERS. After which CNT\_FAM\_MEMBERS was dropped, As \cite{tolosi_classification_2011} a high correlation between features may lead to a biased feature importance ranking and unstable models.

\section{Experiments}

\subsection{Data Pre-processing}

Since the dataset has null values, invalid entries, and duplicates some pre-processing techniques are introduced before it is used for training and testing. 

\paragraph{Null /Empty Entries} Incomplete data is an unavoidable problem in dealing with most real-world data sources \cite{kotsiantis_data_2006}. In this dataset, there is a noticeable amount of null entries. Therefore, features with 30\% of empty entries are removed. 

\paragraph{Invalid features and Duplicate records }
Invalid features are columns in which their information has no meaning for the objective. This includes columns that contain only a single value, or columns that indicate whether a customer provided personal information or not. Hence, the features are dropped.   

From the 36,326 records, we found 25,268 duplicate entries on the dataset. The duplicates were dropped as the presence of duplicates in the learning sample does indeed pose a problem\cite{kolcz_data_2003}.

\textbf{Categorization and outlier removal} An outlier is an unlikely observation in a dataset and may have one of many causes \cite{brownlee_data_2020}. In this case, we removed outlier values of numeric features.

The credit status of the applicants for each month was recorded as c: paid off that month, x: no loan for the month, 0: (1-29) days pass overdue, 1: (30-59) days pass overdue, 2: (60-89) days pass overdue, 3: (90-119) days pass overdue, 4: (120-149) days pass overdue and 5: more than 150 days pass overdue.  We categorized c, x, and 0 as good loans and 1, 2, 3, 4, and 5 as bad loans. 

\textbf{Normalization and encoding} 
The features came in different formats. Eg., string, unbounded integers, floating numbers, Boolean values. This poses a challenge to work with machine learning algorithms. We encoded the categorical features to be represented by integer values. 
One hot encoding is then used for features that are with more than two categories. 

A series distribution of certain data could affect the performance of machine learning \cite{jo_effectiveness_2019}. To get an equal contribution of the features, the numeric features are normalized. After pre-processing, the dataset reduced to 11058 records and 13 features. 

The dataset was categorized into training and testing: 80\% of the dataset for training, 20\% for testing. The imbalance treatment is applied on the training set only. 


\subsection{Imbalance Treatment}
Class imbalance is a problem that exists when the distribution of the target feature has a high variation: one of the target classes is found highly distributed while the other is scarcely distributed.
In the dataset, the ratio of individuals who have entered NPL is very low compared to the total loan. 78.5\% of the data are performing loans while 21.5\% are non performing loans. 
To solve this problem, the following two oversampling techniques are implemented to generate synthetic samples that favour the minority class.


\subsubsection*{ K Means-SMOTE }K Means-SMOTE \cite{last_oversampling_2018}  - works in three steps, the first is to group the input samples into different clusters. The second is to identify a cluster with the low distribution of minority class, assign more synthetic samples, and then finally over-sample each filtered cluster using SMOTE \cite{chawla_smote_2002}. Using this technique we synthesized minority class samples and added them to the original training data.

\subsubsection*{ Conditional Tabular GAN(CTGAN)}
Works based on the principle of the CGAN  model to synthesize new samples that are optimized for tabular data which accounts for continuous and discrete features\cite{xu_modeling_2019}.

\subsection{Important Features  Selection}

It's important to identify a set of significant features relevant to determine the credit risk of the loanee. 
The following methods are applied to get the most important features from the dataset.
 
\textbf{Random Forest feature selection} calculates feature importance based on node impurities at each decision tree and takes a mean of all decision tree feature importance to get the final feature importance. According to \cite{chen_selecting_2020}, RF feature selection method 
are extremely useful and efficient in selecting important features.


\textbf{Extreme Gradient Boosting(XGBoost) }
calculates feature importance after the boosted decision tree is built. \cite{chen_improving_2020} found that XGBoost preserves key features necessary for prediction.

\begin{figure}[ht] 
  \centering
  \subfloat[]{\includegraphics[width=0.5\textwidth]{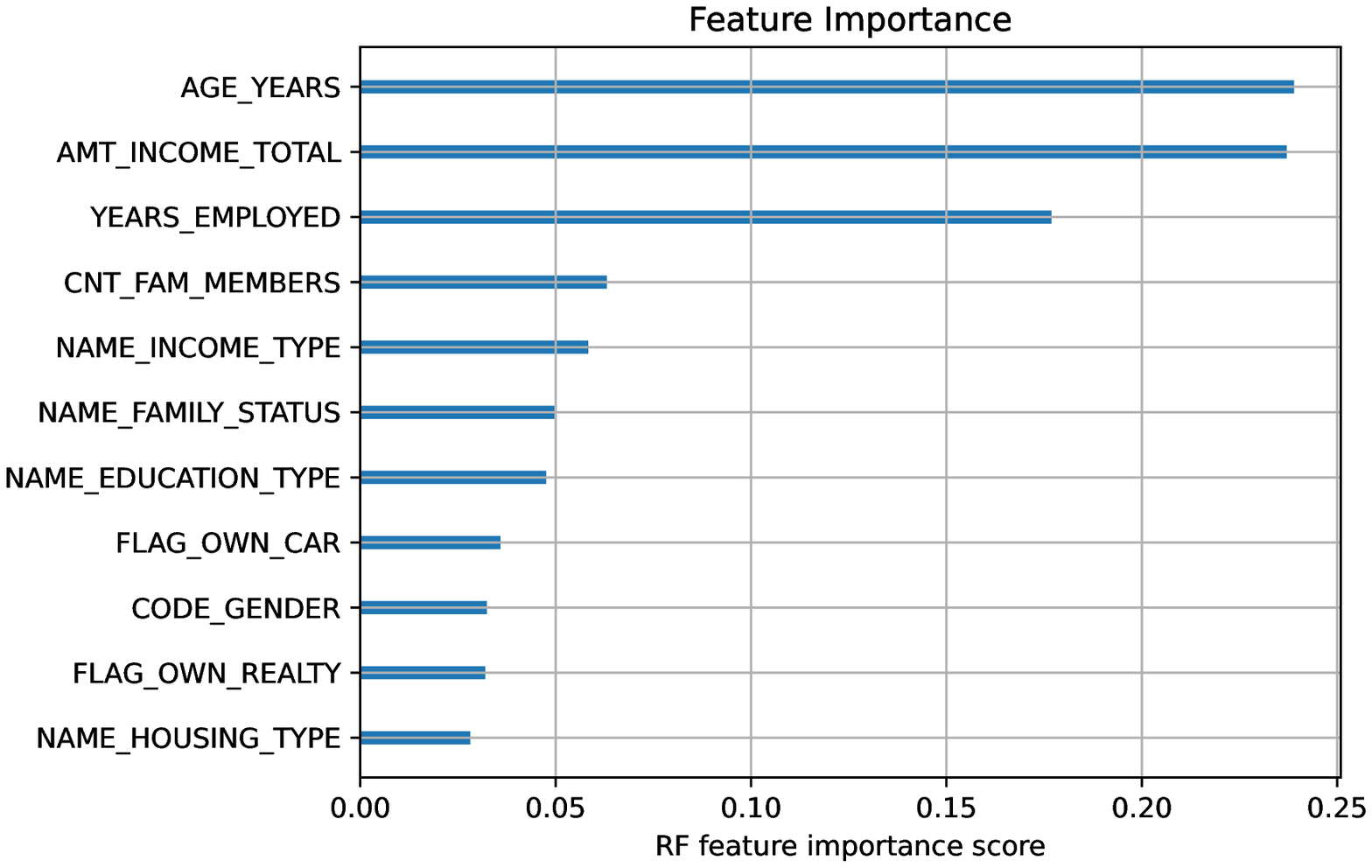}\label{feature-a}}
  \hfill
  \subfloat[]{\includegraphics[width=0.5\textwidth]{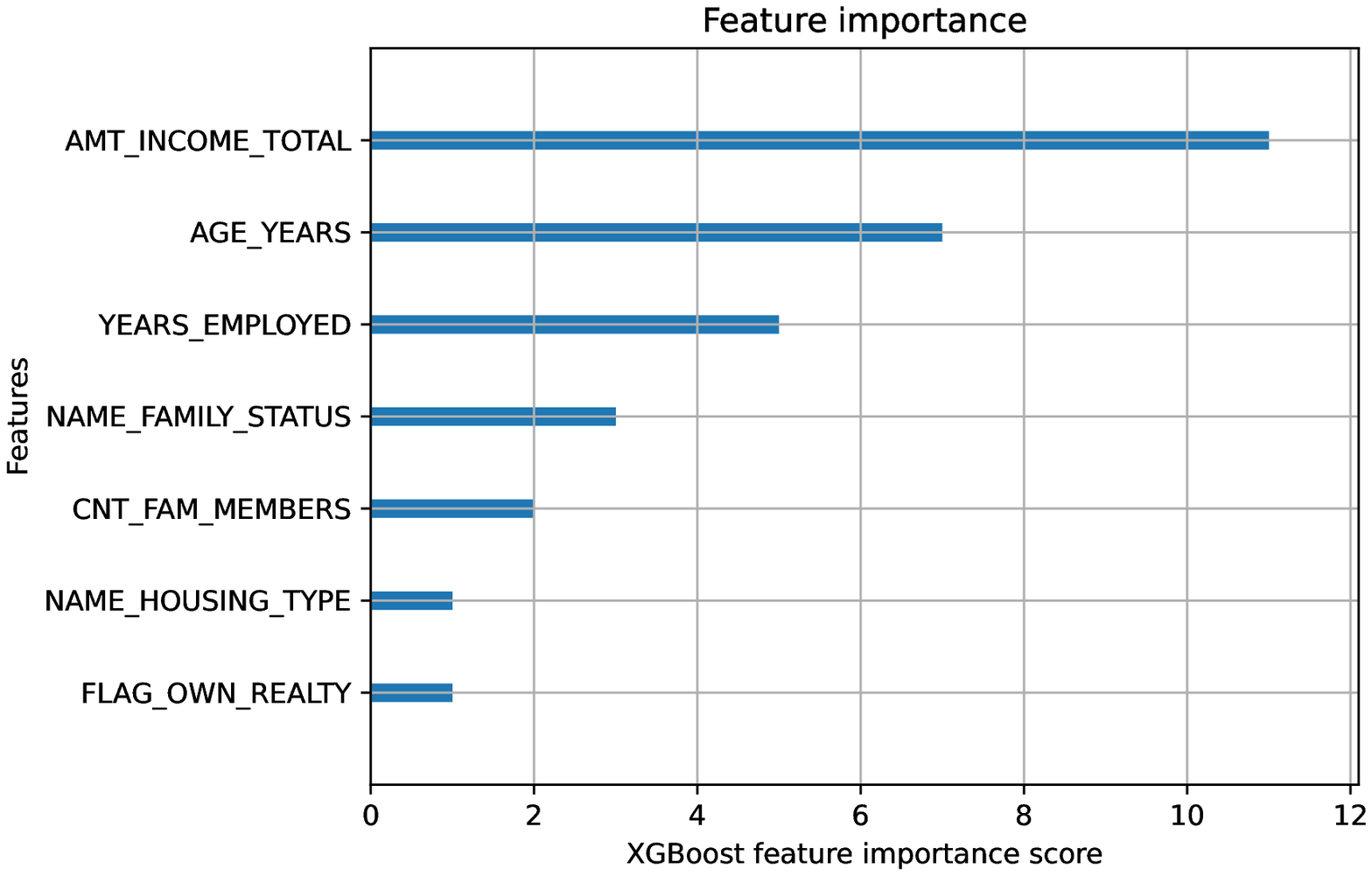}\label{feature-b}}
  \caption{ Feature Importance scores}
  \label{feature importance scores}
\end{figure}

In this work, we have used RF and XGBoost feature selection methods to obtain the feature importance score. The feature importance scores generated by random forest are shown in figure \autoref{feature-a}. While the scores generated by XGBoost are shown in \autoref{feature-b}. The top three important features are selected as there is a high variation compared to the score of the fourth feature. The three important features obtained from both RF and XGBoost are the same. To cross-validate the results of random forest and XGBoost, we implemented Recursive feature elimination using random forest to provide the three most important features which resulted in the same features.
The most important features based on the above experiments are age years, total amount of income, and years of employment.

\section{Results}
In order to evaluate the performance of the models, appropriate metrics should be selected. For the purpose of identifying the non-performing loans we identify positive classes (i.e. high precision) while minimizing those the model is miss-classifying as positive(i.e. high sensitivity). In this case, F1-score, which is the weighted average of Precision and sensitivity, is used as an evaluation metric as it conveys the balance between the precision and the recall. The models are evaluated based on their $F1$ score of the minority class which is $1$.



The results show that the use of K Means-SMOTE over-sampling is effective compared to the unbalanced dataset, under-sampled dataset and a dataset over-sampled using CTGAN. XGBoost achieves the highest performance in terms of F1-Socore.

\begin{table*}[ht] 

\label{table-kmeans-smote}
 \caption{Models performance on KMeans SMOTE over-sampled data}
  \centering
  \resizebox{10cm}{!}{
  \begin{tabular}{llll}
    \toprule
    Model     & Precision(0/1)     & Recall(0/1)  & F1-score(0/1) \\
    \midrule

  Random Forest Classifier  &  0.81/0.90  &  0.91/0.78  &  0.86/0.84   \\ 
  Logistic Regression  &  0.55/0.55  &  0.54/0.56  &  0.55/0.56   \\ 
  Decision Tree Classifier  &  0.79/0.68  &  0.61/0.84  &  0.69/0.75   \\ 
  SVM Classifier  &  0.57/0.56  &  0.51/0.62  &  0.54/0.59   \\ 
  K Neighbors Classifier  &  0.49/0.48  &  0.56/0.41  &  0.52/0.44   \\ 
  XGBoost  &  \textbf{0.87/0.96}  &  \textbf{0.96/0.86}  &  \textbf{0.92/0.91}   \\ 
  Voting Classifier  &  0.86/0.96  &  0.96/0.84  &  0.91/0.90   \\ 
 \bottomrule
  \end{tabular}
  }
  \label{tab:table}
\end{table*}

\begin{table*}[ht] 
 \caption{Models performance on CTGAN over-sampled data}
  \centering
  \resizebox{10cm}{!}{
  \begin{tabular}{llll}
    \toprule
    Model     & Precision(0/1)     & Recall(0/1)  & F1-score(0/1) \\
    \midrule

      Random Forest Classifier  &  \textbf{0.78/0.21}  &  0.95/0.04  &  0.95/0.07   \\ 
  Logistic Regression  &  0.77/0.21  &  0.70/\textbf{0.28}  &  0.70/\textbf{0.24}   \\ 
  Decision Tree Classifier  &  0.77/0.15  &  0.97/0.02  &  0.97/0.04   \\ 
  SVM Classifier  &  0.77/0.20  &  0.74/0.23  &  0.74/0.21   \\ 
  K Neighbors Classifier  &  0.96/0.19  &  0.68/0.26  &  0.68/0.22   \\ 
  XGBoost  &  0.77/0.11  &  \textbf{0.98}/0.01  &  \textbf{0.98}/0.02   \\ 
  Voting Classifier  &  0.77/0.12  &  0.96/0.02  &  0.96/0.03   \\ 
 \bottomrule
  \end{tabular}
  }
  \label{tab:table}
\end{table*}


\section{Conclusion}
In this work, we evaluated the performance of various machine learning algorithms on predicting non-performing loans. We introduced different imbalance treatment techniques like CTGAN and kmeans-SMOTE on our dataset. The model's performance was evaluated on unbalanced, under-sampled, over-sampled (using CTGAN and Kmeans-SMOTE) dataset. \newline
The results shows that the kmeans-SMOTE over-sampling technique improved the models performance. XGBoost on the kmeans-SMOTE over-sampled data achieved the highest performance in terms of F1-Score with a very high
margin as compard to CTGAN. \newline
We also selected the most important features using different feature selection methods on the dataset. The results show that Age years, total amount of income, and years employed are the most determining features. This indicates that banks should give attention to those features in addition to the current working system.  
For future works, we will work on enhanced feature selection methods. We encourage researchers to explore more towards such methodologies to predict defaulting loans.  

\section{Data Availability}

The dataset used and analysed during the current study is not publicly available due to the  bank's principle, but can be available from the corresponding author on reasonable request.

\bibliographystyle{unsrtnat}  
\bibliography{references}

\begin{thebibliography}{21}
\providecommand{\natexlab}[1]{#1}
\providecommand{\url}[1]{\texttt{#1}}
\expandafter\ifx\csname urlstyle\endcsname\relax
  \providecommand{\doi}[1]{doi: #1}\else
  \providecommand{\doi}{doi: \begingroup \urlstyle{rm}\Url}\fi

\bibitem[Adhikary()]{adhikary_nonperforming_nodate}
Bishnu~Kumar Adhikary.
\newblock Nonperforming loans in the banking sector of bangladesh: Realities
  and challenges.
\newblock page~21.

\bibitem[Li et~al.(2010)Li, Wang, and Wang]{li_comparison_2010}
Feng-Chia Li, Peng-Kai Wang, and Gwo-En Wang.
\newblock Comparison of the primitive classifiers with extreme learning machine
  in credit scoring.
\newblock pages 685--688, 2010.
\newblock \doi{10.1109/IEEM.2009.5373241}.

\bibitem[Heath(2019)]{heath_prediction_2019}
Donald~R. Heath.
\newblock Prediction machines: the simple economics of artificial intelligence.
\newblock 21\penalty0 (3):\penalty0 163--166, 2019.
\newblock ISSN 1522-8053.
\newblock \doi{10.1080/15228053.2019.1673511}.
\newblock URL \url{https://doi.org/10.1080/15228053.2019.1673511}.
\newblock Publisher: Routledge \_eprint:
  https://doi.org/10.1080/15228053.2019.1673511.

\bibitem[Zhu et~al.(2019)Zhu, Qiu, Ergu, Ying, and Liu]{zhu_study_2019}
Lin Zhu, Dafeng Qiu, Daji Ergu, Cai Ying, and Kuiyi Liu.
\newblock A study on predicting loan default based on the random forest
  algorithm.
\newblock 162:\penalty0 503--513, 2019.
\newblock \doi{10.1016/j.procs.2019.12.017}.

\bibitem[Ghatasheh(2014)]{ghatasheh_business_2014}
Nazeeh Ghatasheh.
\newblock Business analytics using random forest trees for credit risk
  prediction: A comparison study.
\newblock 72:\penalty0 19--30, 2014.
\newblock \doi{10.14257/ijast.2014.72.02}.

\bibitem[Breeden(2020)]{breeden_survey_2020}
Joseph Breeden.
\newblock \emph{A Survey of Machine Learning in Credit Risk}.
\newblock 2020.
\newblock \doi{10.13140/RG.2.2.14520.37121}.

\bibitem[Li(2019)]{li_credit_2019}
Yu~Li.
\newblock Credit risk prediction based on machine learning methods.
\newblock In \emph{2019 14th International Conference on Computer Science
  Education ({ICCSE})}, pages 1011--1013, 2019.
\newblock \doi{10.1109/ICCSE.2019.8845444}.
\newblock {ISSN}: 2473-9464.

\bibitem[Pradhan et~al.(2020)Pradhan, Akter, and
  Marouf]{pradhan_performance_2020}
Mohammad Pradhan, Sima Akter, and Ahmed Marouf.
\newblock Performance evaluation of traditional classifiers on prediction of
  credit recovery.
\newblock pages 541--551. 2020.
\newblock ISBN 9789811555572.
\newblock \doi{10.1007/978-981-15-5558-9_48}.

\bibitem[Granström and Abrahamsson(2019)]{granstrom_loan_2019}
Daria Granström and J.~Abrahamsson.
\newblock Loan default prediction using supervised machine learning algorithms.
\newblock 2019.
\newblock URL
  \url{https://www.semanticscholar.org/paper/Loan-Default-Prediction-using-Supervised-Machine-Granstr%C3%B6m-Abrahamsson/39a023b32bfae8f39f428b29c95d0ae3b9191114}.

\bibitem[Chen and Zhang(2021)]{chen_research_2021}
Ying Chen and Ruirui Zhang.
\newblock Research on credit card default prediction based on k-means {SMOTE}
  and {BP} neural network.
\newblock 2021:\penalty0 e6618841, 2021.
\newblock ISSN 1076-2787.
\newblock \doi{10.1155/2021/6618841}.
\newblock URL \url{https://www.hindawi.com/journals/complexity/2021/6618841/}.
\newblock Publisher: Hindawi.

\bibitem[Gültekin and Erdoğdu~Şakar(2018)]{gultekin_variable_2018}
Başak Gültekin and Betül Erdoğdu~Şakar.
\newblock Variable importance analysis in default prediction using machine
  learning techniques:.
\newblock In \emph{Proceedings of the 7th International Conference on Data
  Science, Technology and Applications}, pages 56--62. {SCITEPRESS} - Science
  and Technology Publications, 2018.
\newblock ISBN 978-989-758-318-6.
\newblock \doi{10.5220/0006872400560062}.
\newblock URL
  \url{http://www.scitepress.org/DigitalLibrary/Link.aspx?doi=10.5220/0006872400560062}.

\bibitem[Toloşi and Lengauer(2011)]{tolosi_classification_2011}
Laura Toloşi and Thomas Lengauer.
\newblock Classification with correlated features: unreliability of feature
  ranking and solutions.
\newblock 27\penalty0 (14):\penalty0 1986--1994, 2011.
\newblock ISSN 1367-4803.
\newblock \doi{10.1093/bioinformatics/btr300}.
\newblock URL \url{https://doi.org/10.1093/bioinformatics/btr300}.

\bibitem[Kotsiantis et~al.(2006)Kotsiantis, Kanellopoulos, and
  Pintelas]{kotsiantis_data_2006}
Sotiris Kotsiantis, Dimitris Kanellopoulos, and P.~Pintelas.
\newblock Data preprocessing for supervised learning.
\newblock 1:\penalty0 111--117, 2006.

\bibitem[Kołcz et~al.(2003)Kołcz, Org, {Chowdhury}, and
  Alspector]{kolcz_data_2003}
Aleksander Kołcz, Abdur Org, {Chowdhury}, and Joshua Alspector.
\newblock Data duplication: an imbalance problem?
\newblock 2003.

\bibitem[Brownlee(2020)]{brownlee_data_2020}
Jason Brownlee.
\newblock Data preparation for machine learning: Data cleaning, feature
  selection, and data transforms in python.
\newblock pages 54 -- 65. 2020.
\newblock
  \doi{https://books.google.com.et/books/about/Data_Preparation_for_Machine_Learning.html?id=uAPuDwAAQBAJ&redir_esc=y}.

\bibitem[Jo(2019)]{jo_effectiveness_2019}
Jun-Mo Jo.
\newblock Effectiveness of normalization pre-processing of big data to the
  machine learning performance.
\newblock 14\penalty0 (3):\penalty0 547--552, 2019.
\newblock ISSN 1975-8170.
\newblock \doi{10.13067/JKIECS.2019.14.3.547}.
\newblock URL
  \url{https://www.koreascience.or.kr/article/JAKO201924763903550.page}.
\newblock Publisher: Korea Institute of Electronic Communication Science.

\bibitem[Last et~al.(2018)Last, Douzas, and Bacao]{last_oversampling_2018}
Felix Last, Georgios Douzas, and Fernando Bacao.
\newblock Oversampling for imbalanced learning based on k-means and {SMOTE}.
\newblock 465:\penalty0 1--20, 2018.
\newblock ISSN 00200255.
\newblock \doi{10.1016/j.ins.2018.06.056}.
\newblock URL \url{http://arxiv.org/abs/1711.00837}.

\bibitem[Chawla et~al.(2002)Chawla, Bowyer, Hall, and
  Kegelmeyer]{chawla_smote_2002}
Nitesh Chawla, Kevin Bowyer, Lawrence Hall, and W.~Kegelmeyer.
\newblock {SMOTE}: Synthetic minority over-sampling technique.
\newblock 16:\penalty0 321--357, 2002.
\newblock \doi{10.1613/jair.953}.

\bibitem[Xu et~al.(2019)Xu, Skoularidou, Cuesta-Infante, and
  Veeramachaneni]{xu_modeling_2019}
Lei Xu, Maria Skoularidou, Alfredo Cuesta-Infante, and Kalyan Veeramachaneni.
\newblock Modeling tabular data using conditional {GAN}.
\newblock 2019.
\newblock URL \url{http://arxiv.org/abs/1907.00503}.

\bibitem[Chen et~al.(2020{\natexlab{a}})Chen, Dewi, Huang, and
  Caraka]{chen_selecting_2020}
Rung-Ching Chen, Christine Dewi, Su-Wen Huang, and Rezzy~Eko Caraka.
\newblock Selecting critical features for data classification based on machine
  learning methods.
\newblock 7\penalty0 (1):\penalty0 52, 2020{\natexlab{a}}.
\newblock ISSN 2196-1115.
\newblock \doi{10.1186/s40537-020-00327-4}.
\newblock URL \url{https://doi.org/10.1186/s40537-020-00327-4}.

\bibitem[Chen et~al.(2020{\natexlab{b}})Chen, Zhang, Yu, Yu, Lawrence, Ma, and
  Zhang]{chen_improving_2020}
Cheng Chen, Qingmei Zhang, Bin Yu, Zhaomin Yu, Patrick~J. Lawrence, Qin Ma, and
  Yan Zhang.
\newblock Improving protein-protein interactions prediction accuracy using
  {XGBoost} feature selection and stacked ensemble classifier.
\newblock 123:\penalty0 103899, 2020{\natexlab{b}}.
\newblock ISSN 0010-4825.
\newblock \doi{10.1016/j.compbiomed.2020.103899}.
\newblock URL
  \url{https://www.sciencedirect.com/science/article/pii/S0010482520302481}.

\end{thebibliography}

\end{document}